# Optimization of side lobe level of linear antenna array using nature optimized ants bridging solutions(NOABS)

Sunit Shantanu Digamber Fulari
Dept. of electronics and communication *(Chandigarh University)*
Mohali, India

*Abstract*—Nature inspired algorithms has brought solutions to complex problems in optimization where the optimization and solution of complex problems is highly complex and nonlinear. There is a need to use proper design of the cost function or the fitness function in terms of the parameters to be optimized, this can be used in solving any type of such problems. In this paper the nature inspired algorithms has played important role in the optimal design of antenna array with improved radiation characteristics. In this paper, 20 elements linearly spaced array is used as an example of nature inspired optimization in antenna array system. This bridge inspired army ant algorithm(NOABS) is used to reduce the side lobes and to improve the other radiation characteristics to show the effect of the optimization on design characteristics by implementation of NOABS nature inspired algorithm. The entire simulation is carried out on 20 elements linear antenna array.

## I. INTRODUCTION (*ANT BRIDGES*)

Self-organization of the ants is remarkable which help them to survive in times of adversity. They maintain a close proximity with each other when there is a predator attach by which they leave out chemicals and pheromones into their paths which triggers a panic attack. The ants are involved in building of their ant hills, they are built in turn by the Formica and carpenter ants by burrowing in mud and then depositing it in the form of ant hill. The fitness of the ant is taken into account; in addition the strength of the ant hill is also important. The ant hill is a unique structure which involves individual departments allowing the ants for rest, storing their food and eggs. The ant hill consists of multiple departments or inner structural layers which help in multiple tasks. The ants store their eggs at the top for warmness during the say while they shift it to bottom during the night. The nature inspired optimized ants bridging solution(NOABS) is a nature inspired algorithm used in this paper to synthesize the antenna for side lobe reduction and finding the null depths in linear antenna array. In this simulation it is observed that in 20 element linear array we have successfully synthesized the antenna to find the null depths, the side lobes have been synthesized reduced and further the fitness value of the function is taken into consideration in terms of the army ants bridge to optimize the antenna. The desired null is shown in figure VII.

## II Concept:

Figure II: Bridge between leaves

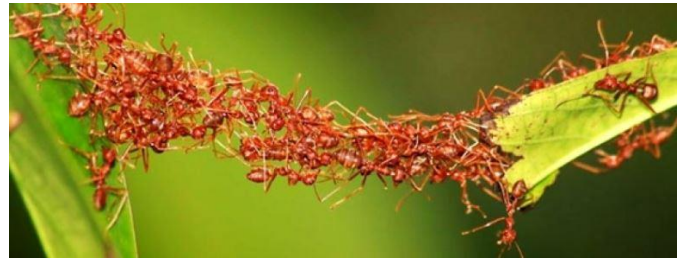

The ants when there is a gap or obstruction in their way want to build a cross connection in the air, they kind of wind up together to allow the preceding ants to pass over them. This helps in building the bridge channel between themselves for allowing the ants to pass over. The fitness value of the ants is taken into custody while dealing with ant hill bridges. The strength of the ants in holding the entire bridge is very crucial in this regards. There is a propositional fact about these ants that they can hold 100 times their weight or can carry one hundredth times their weight, this makes them very strong in building transplanted bridges in the air holding two separated structures apart. But not all ant bridges are so strong which can hold, some are weak and can also lead to collapse due to external factors and air pressure.

## III Novelty:

We are following with a random experiment in which the chance of the ant bridge to stand and also fall is taken into account. There is nature selection process taken into account, when will the ant bridge fall or break. We start guessing randomly, then realizing that the experiments are too much dependent on chance. When guessing we take into account the fitness value of the ants, how strong or fitness value the ant is while holding the entire folk together. Then we look for variations to find out if there are more bridges of ants or which can be bettered in some way. There is a different psychology of the ants while building the bridge. Considering a path which is widening out, when selecting where to build the ant bridge, the ants don't select the shortest or bridge which makes the linkage the most efficient but does select a bridge where they have a kind of psychology in their setup. The army ants does not always take the bridge construction which makes the shortest path, they are more having engineering sense as they choose a longer path bridge as it makes it more safer for travelling, this is the sense of engineering in army ants.

Figure III: Army ant bridging technique

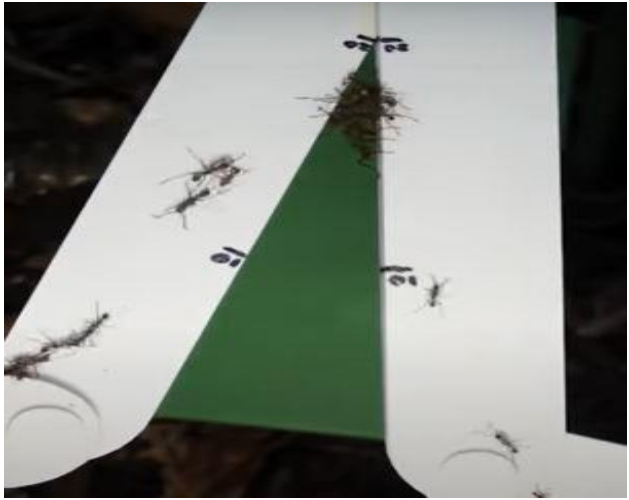

The ant bridges are done by army ants. There is some other calculation in their unconscious psychology. Scientists are working on swarm of ants robots who try to establish similar bridges artificially. Bridges are formed by interconnections between the ants and how strongly the ants can hold each other to formation of the bridge. These factors determine how the bridge can be held firmly.

## IV Proposed Work:

Cost metric framework:

Considering there is no bridge than the additional distance the ants must travel is given by the following equation.

$$\frac{Number\ of\ ants}{Total\ distance} = \frac{N}{L_T + L_A} \quad \text{--(1)}$$

The magnitude of the length $L_A$ depends on the assumption and configuration that the ants try to travel through the inner edge of the trail to minimize the travel distance. The trail length $L_T$ is fixed. The construction of the architectural bridges modifies the number of ants travelling through the path leaving behind chemical and pheromones but also modifies the total distance of the overall path the ants travel drastically.

The total number of ants which follow the process become N minus the number of ants sequential zed in the bridge structure.

$$N - n_b \quad \text{--(2)}$$

The foraging ants have a typical difference between the real bridges built and the structurally built bridges by the army ants. There is a thorough research done on the type of ants which have a role in building these ant bridges. It is studied that these ants building the bridges tend to be little less in size as compared to the other ants as they can easily structure the bridges, besides they are fundamentally less effective in carrying food and prey items. So considering these equations the number of foraging ants in very less compared to the number of bridge formation ants. Therefore due to the difference in the foraging ants and ant building bridge ants we introduce an coefficient alpha $\alpha$ known as below equation

$$N - \frac{nb}{\alpha} \quad \text{--(3)}$$

We set $\alpha = 17.02$ as stated by Reid et.al. The value of this coefficient needs to be refit when testing for the foraging and bridge building ants during the nighttime when there are structural fundamental difference between the ants.

$$\rho = \frac{N - \frac{nb}{\alpha}}{f} \quad \text{--(4)}$$

The above equation is on the shortest distance optimization function to be optimized when constructing the bridge. f being the shortest distance in the presence of the bridge, which is the shortest distance in the presence of the foraging bridge.

Forces equations considering the army ant bridge:

$\sum Fx = 0$ This is the force in the right hand direction.

$\sum Fy = 0$ This is the forces in the upward direction.

$\sum Ma = 0$ This is the moment of force in the counter clockwise direction.

Moment of force is $F * \Delta T$ or force times perpendicular distance where the force is acting given in the below figure.

The force is inclined towards the centre of gravity of the ant system. This is inturn made balanced by the ant colony system. These equations are made to implement in this simulation.

The ant system is a perfect example of working in unity in the ant species system. Without inclining towards each other the ants bridge is impossible, this is in turn impossible if the AHCOA algorithm and working was not in tune with the ant bridging system. The AHCOA and NOABS is working in harmony with each other.

The ant colony system functions with unity with each other, where the entire world is totally interdependent on the functioning and helping hand of each other. We have simulated for this algorithm the antenna pattern synthesis.

Figure IV:

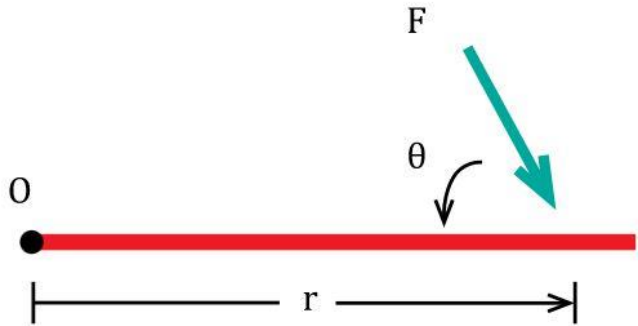

There is a kind of technique employed by ants in which the ants leave slightly small gap between themselves when they build the ant bridge, one ant leaves slightly small gap between the next when building the bridge, consider the human bridge. It is not always possible to build a standing bridge in the air without pillars as built by the ants. But the ants very constructively build their ant bridges without much fault and flaws. This is a classic example of mechanics which is used by civil engineers in the design of the bridges. The ants also implement this similar concept in the design of their bridges, it is a matter of chance the survival of the bridge.

Figure V:

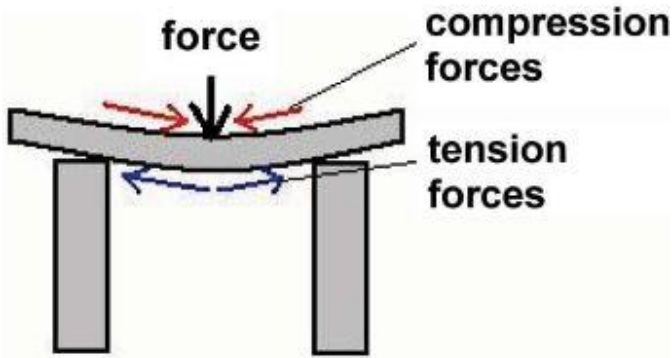

The forces acting on the bridge considering normal bridge, it may be ant bridge or human constructed bridge is the tension forces and the compression forces acting on the bridge structure.

### V Related work:

In a paper by Rahmat Samii [1] particle swarm optimization a nature inspired algorithm is used to optimize antenna. In earlier methods genetic algorithm was also implemented in antenna synthesis optimization. Saxena et.al in [2] has used lion optimizer in antenna side lobe reduction.

### VI Experiments and Results:

Figure VI:

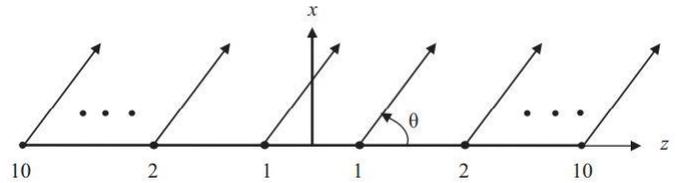

Consider a 20 element linearly spaced array as arranged as follows.

In the first step we optimize the array excitation magnitudes of the array elements so that the corresponding array factor(AF) has nulls at specific directions.

For a 20 spaced linear array, The AF can be written as

$$AF(\theta) = 2\sum_{n=1}^{10} a(n) e^{j\varphi(n)} \cos[\beta d(n)\cos(\theta)]$$

Where $\beta$ is the wave number, $a(n)$, $\varphi(n)$ and $d(n)$ are the excitation magnitude, phase and location of the $n^{th}$ element respectively. Out of 20 we need to optimize half or 10 linear array elements which gives us the equation as(phase element becomes 0 or zero)

$$AF(\theta) = 2\sum_{n=1}^{10} a(n) \cos[\beta d(n)\cos(\theta)]$$

The fitness function is given by

Fitness value of function in optimization=

$$\int_0^{180} [AF(\theta - AFd(\theta)] \left[\frac{1+sgn(AF(\theta)-AFd(\theta))}{2}\right] d\theta$$

Figure VII:

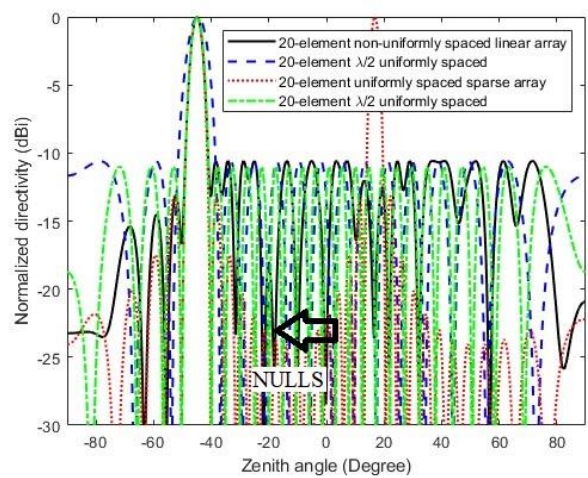

Figure VIII: Side lobes and desired nulls schematic.

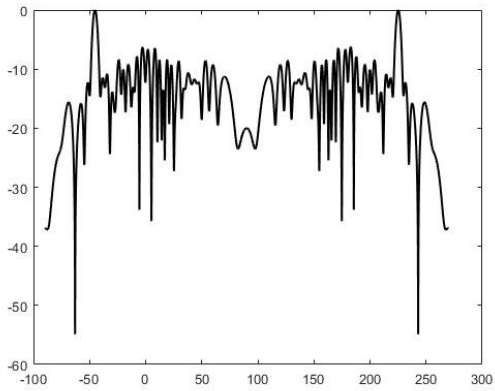

The following graphs are for 20 element equally spaced linear array. The graph shows a peculiar variations for the army ants bridge problem antenna side lobe reduction methodology.

Figure X: NOABS algorithm lobes

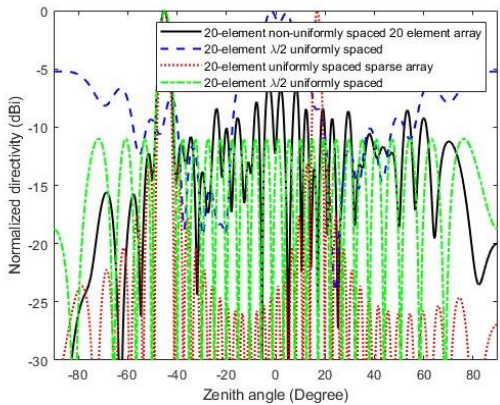

Figure IX: Pictographic representation of side lobes

Extracted plot

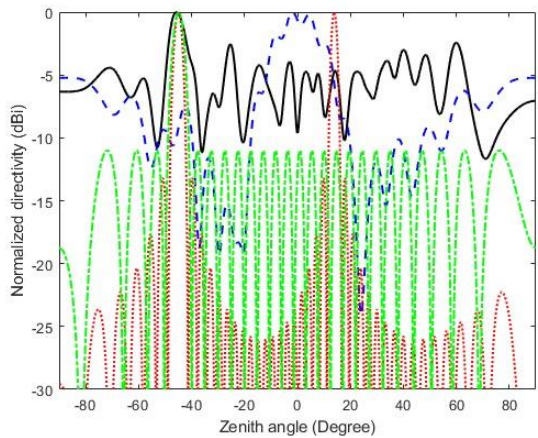

Figure X: 20 element shifted graph extracted plot

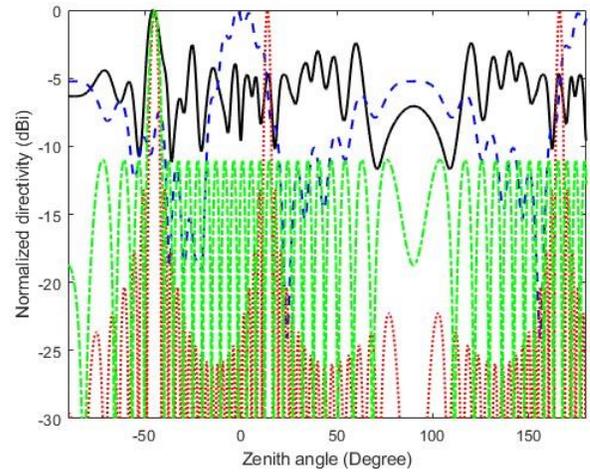

Figure XI: Normal dense array value shifted from 0.05 to 0.1.

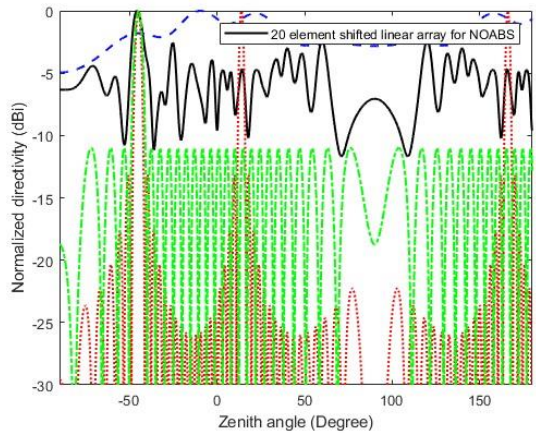

The above figure shows the radiation pattern with side lobe levels and null depths simulated. The NOAB shows reduced side lobe levels as compared to other state of the art nature inspired algorithms such as ant lion optimizer and particle swarm optimization, including genetic algorithm. The army ants bridge is a chance of standing in times when there is a need for linkage between paths. Depending on the fitness value of the ants and overall bridge the bridge survives.

**VII Conclusion**

The NOABS algorithm, is a new nature inspired algorithm, introduced to the electromagnetic and antenna community in this letter. This algorithm showed improvement as compared to the previous AHCOA, ant lion optimizers, genetic algorithm and also particle swarm optimization. On the basis of the antenna a 20 element antenna array is designed. The NOABS algorithm showed improved null depths and lower side lobe levels as compared to previously state of the art metaheuristic algorithms used in similar optimizations. Furthermore, the NOABS algorithm showed high efficiency and better radiation simulation pattern.